\documentclass[a4paper]{article}

\usepackage{fancyhdr}
\usepackage{comment}
\usepackage{amsmath,amssymb}
\usepackage{bm}
\usepackage{graphicx}
\usepackage[dvipdfmx]{color}
\usepackage{ascmac}
\usepackage{url}
\usepackage{siunitx}
\usepackage{float}
\usepackage{slashbox}
\usepackage{here}
\usepackage{txfonts}
\usepackage{listings}
\usepackage{tikz}

\newcolumntype{G}{>{\columncolor[gray]{0.8}}c}
\newcommand{\hcircle}[1]{\mathrm{\scriptsize \textcircled{\tiny \raisebox{-0.5pt}[0ex][0ex]{H}}}}

\newcommand{\vSet}[1]{V_{#1}}

\newcommand{\infofl}[1]{\mathit{Info}({#1})}
\newcommand{\infofr}[1]{-\sum_{y\in Y}\frac{|{#1^y}|}{|{#1}|}\log_2\frac{|{#1^y}|}{|{#1}|}}

\newcommand{\ainfofu}[2]{\mathit{Info}_{#1}({#2})}
\newcommand{\myainfofr}[3]{\sum_{x\in \{{#1},-{#1}\}}\frac{|{#3^x}|}{|{#2}|}\infofl{#3^x}}

\newcommand{\mysinfofr}[3]{-\sum_{x\in \{{#1},-{#1}\}}\frac{|{#3^x}|}{|{#2}|}\log_2\frac{|{#3^x}|}{|{#2}|}}

\newcommand{\myigrfl}[2]{\mathit{IGR}_{#1}({#2})}

\title{Multi-duplicated Characterization of Graph Structures using Information Gain Ratio for Graph Neural Networks} 
\author{Yuga Oishi\footnotemark[1] \;\;\;\;\;Ken Kaneiwa\footnotemark[1]}
\date{\footnotesize	\footnotemark[1] Department of Computer and Network Engineering, Graduate School of Informatics and Engineering,
The University of Electro-Communications, Tokyo, Japan
\\
y.oishi@sw.cei.uec.ac.jp}

\begin{document}
\maketitle

\begin{abstract}
Various graph neural networks (GNNs) have been proposed to solve node classification tasks in machine learning for graph data. 
GNNs use the structural information of graph data by aggregating the features of neighboring nodes.
However, they fail to directly characterize and leverage the structural information.
In this paper, we propose multi-duplicated characterization of graph structures using information gain ratio (IGR) for GNNs (MSI-GNN), which enhances the performance of node classification by using an i-hop adjacency matrix as the structural information of the graph data.
In MSI-GNN, the i-hop adjacency matrix is adaptively adjusted by two methods: (i) structural features in the matrix are selected based on the IGR, and (ii) the selected features in (i) for each node are duplicated and combined flexibly.
In an experiment, we show that our MSI-GNN outperforms GCN, H2GCN, and GCNII in terms of average accuracies in benchmark graph datasets.
\end{abstract}

\section{Introduction}
Various real-world networks, such as citation networks of papers and link relations of web-pages, can be expressed as graph data.
In recent years, there have been many explorations on machine learning for graph data mainly for link prediction \cite{transe, rgcn, compgcn}, graph classification \cite{graphclass}, and node classification \cite{node2vec, lp, minami}.
In particular, graph convolutional networks (GCNs) \cite{gcn} have been attracting attention for node classification because of their high accuracies compared to those of conventional methods.
Subsequently, various graph neural networks (GNNs) \cite{gnn, gin, gat, jknet, gcnii} based on GCN have been proposed to improve these accuracies.
GNNs can be applied to a wide range of fields, such as recommendation systems \cite{recom}, traffic prediction \cite{astgcn, gcnhlstm}, natural language processing \cite{gcnfornlp}, and computer vision \cite{gcncompvision}.

%そもそもいらない？
%There is a technique called graph embedding to solve the node classification.
%Graph embedding encodes the structural information of graph data and applies the generated vector to SVM and so on.
%As prior research, node2vec\cite{node2vec} combines depth-first search and breadth-first search, and GraphSAGE\cite{sage} learns aggregate functions that can also generate unknown node embeddings. 

A GNN learns the label information of nodes in the graph by aggregating the feature vectors of their neighboring nodes.
The GNN captures the structural information of graph data by extracting only the information of neighboring nodes in the aggregation function.
In other words, the quality of the aggregation function and feature vectors is crucial to node classification in a GNN.

Some studies on GNN have attempted to make greater use of the structural information of graph data.
For example, Geom-GCN \cite{geom} extracts the structural information of neighboring nodes using geometric relationships in the latent space.
The structural properties of graph data also include concepts such as homophily and heterophily \cite{heterosurvey}.
Nodes with the same label tend to be adjacent in a graph that exhibits homophily, whereas nodes with different labels tend to be adjacent in a graph that exhibits heterophily.
Unfortunately, conventional GNNs tend not to perform well on graph data that exhibit heterophily.
To improve the accuracy on such graphs, H2GCN \cite{h2} distinguishes aggregations by depth, whereas frequency adaptation GCN (FAGCN) \cite{fagcn} extracts information flexibly by combining high-pass and low-pass filters.
These GNNs are effective for the node classification of various graph data.
However, they utilize the structural information of graph data only in aggregation functions and do not utilize that in feature vectors.

In this paper, we propose multi-duplicated characterization of graph structures using information gain ratio (IGR) for GNNs (MSI-GNN), which characterizes the structural information of graph data and improves the effectiveness of conventional GNNs.
In MSI-GNN, the structural information of graph data is extracted using an $i$-hop adjacency matrix that expresses the connectivity with neighboring nodes in $i$-hop destinations.
Only the useful structural features are then selected for node classification using the information gain ratio and occurrence filter.
MSI-GNN also duplicates some vectors constructed by the useful features and adaptively combines them.
As a result, MSI-GNN can adjust the ratio of vectors based on the usefulness of each vector and use it as input.
MSI-GNNs can be applied to various conventional GNNs.
The contributions of this study are summarized as follows.
\begin{itemize}
\item MSI-GNN characterizes the structural information of graph data, which is otherwise not fully utilized in conventional GNNs, and can be applied to various GNNs.
\item A method for selecting only the useful structural information for node classification using the information gain ratio and occurrence filter is developed. Furthermore, flexible and powerful learning is realized via the definition of MSI-layers that adaptively duplicate and combine the features.
\item In evaluation experiments using various datasets, we prove that MSI-GNN achie\-ves higher accuracies than those of conventional models and analyze the effectiveness of parameters specific to MSI-GNN.
\end{itemize}	
	
	The rest of the paper is organized as follows.
	In Section 2, we describe the node classification of graph data and GNNs.
	In Section 3, we provide an analysis of the node classification of graph data.
	In Section 4, we define MSI-GNN, which characterizes structural information, selects only useful features, and adaptively duplicates and combines features. Furthermore, we demonstrate how to apply MSI-GNNs to conventional GNNs.
	In Section 5, we provide a review of related studies.
	In Section 6, we recount the evaluation experiments and present an analysis of MSI-GNN using multiple datasets.
	Finally, in Section 7, we conclude the paper and discuss future work.

\section{Notation and Preliminaries}

\subsection{Semi-Supervised Learning for Node Classification of Graph Data}
We describe semi-supervised learning for the node-classification problem on graph data. Let $G = (V, E)$ be a graph with a node set $V$ and an edge set $E$. Each node $v \in V_L$ contained in a subset of the node set $V_L \subset V$ has a label $y_v \in Y$ where $Y$ is a label set. The objective of node classification is to predict the labels of unlabeled nodes $v \in V_U = V \backslash V_L$. We can use an $F$-dimensional feature vector $X_v\in \mathbb{R}^F$ for each node $v\in V$ for learning.

\subsection{Graph Neural Networks}\noindent
\textbf{GNN} : A graph neural network (GNN) \cite{gnn} is a framework of machine learning methods that learns label information by aggregating feature vectors of neighboring nodes, and is generally expressed by the following equation:
\begin{eqnarray}
h_{v}^{(k)} &= & f \Bigl( {h}_{v}^{(k - 1)}, \{ {h}_{u}^{(k - 1)} : u \in N_i(v) \} \Bigr), \nonumber
\end{eqnarray}
where $h_{v}^{(0)} = X_v$ and $N_i(v)$ are neighboring nodes located $i$-hops away from node $v$.
Additionally, the function $f$ is a model-specific function and is applied to aggregate the information of neighboring nodes in each layer.\\

\noindent
\textbf{GCN} : GCN \cite{gcn}, the most basic GNN model, learns weights by aggregating the feature vectors of neighboring nodes and averaging them.
The output $h_{v}^{(k)}$ of node $v$ in the $k$-th layer of a GCN is expressed by the following equations: 
\begin{eqnarray}
{h'}_{v}^{(k)} & = &\sum_{u \in N_{1}(v) \cup \{v\}} h_{u}^{(k-1)} (d_{1,v}+1)^{-1/2}(d_{1,u}+1)^{-1/2},  \nonumber \\
h_{v}^{(k)} &= & \sigma \Bigl( {h'}_{v}^{(k)} W_k \Bigr), \nonumber
\end{eqnarray}
where $h_{v}^{(0)} = X_v$, $d_{i,v} = |N_{i}(v)|$, $W_k$ is the learnable weight matrix for the $k$-th layer, and $ \sigma$ is the activation function.\\

\noindent
\textbf{GCNII} : GCNII \cite{gcnii} is a GNN model that prevents decreases in the accuracy caused by the over-smoothing that occurs when the number of GNN layers is increased.
GCNII uses an initial residual connection to leave the representation in the input layer in all hidden layers, and identity mapping to reduce information loss.
The output $h_v^{(k)}$ of node $v$ in the $k$-th layer of a GCNII is expressed by the following equations:
\begin{eqnarray}
h_{v}^{(0)} & = & \sigma \bigl( X_v W_0 + b \bigr), \nonumber\\
{h'}_{v}^{(k)} & = & (1 - \alpha_{k-1}) \Bigl( \sum_{u \in N_{1}(v) \cup \{v\}} h_{u}^{(k-1)}(d_{1,v}+1)^{-1/2}(d_{1,u}+1)^{-1/2} \Bigr) + \alpha_{k-1} h_{v}^{(0)},  \nonumber \\
h_{v}^{(k)} &= & \sigma \Biggl( {h'}_{v}^{(k)} \Bigl( (1 - \beta_{k-1}) I_{n} + \beta_{k-1} W_k \Bigr) \Biggr), \nonumber
\end{eqnarray}
where $b$ is the learnable bias weight, $I_n$ is the identity matrix, and $\alpha$ and $\beta$ are the hyperparameters for the initial residual connection and identity mapping, respectively.\\

\noindent
\textbf{H2GCN} : H2GCN \cite{h2} is a GNN model that enhances learning ability by distinguishing the aggregated feature vectors by depth.
The 0-th layer $h_v^{(0)}$: the representation $h_{v,i}^{(k)}$, which aggregates the vectors of nodes at depth $i$ from node $v$ in the $k$-th layer: the hidden layer $h_v^{(k)}\ (1 \leq k \leq K)$: and the output layer $h_v^{(\mathrm{output})}$ are expressed by the following equations:
\begin{eqnarray}
h_{v}^{(0)} & = & \sigma \bigl( X_v\ W_e \bigr), \nonumber\\
h_{v,i}^{(k)} & = & \sum_{u \in N_{i}(v)} h_{u}^{(k-1)} d_{i,v}^{-1/2}d_{i,u}^{-1/2}, \nonumber\\
h_{v}^{(k)} & = & \mathrm{CONCAT} \bigl( h_{v,1}^{(k)}, h_{v,2}^{(k)} \bigr), \nonumber\\
h_{v}^{(\mathrm{output})} & = & \mathrm{softmax} \bigl( \mathrm{CONCAT} \bigl( h_{v}^{(0)}, h_{v}^{(1)}, ..., h_{v}^{(K)} \bigr) W_c \bigr), \nonumber
\end{eqnarray}
where $W_e$ and $W_c$ are learnable weight matrices.

\section{Analysis of Node Classification of Graph Data}
\subsection{Usefulness of Structural Information}
\begin{figure}[H]
\centering
%\hspace{-1.0cm}
  \includegraphics[width=11.0cm]{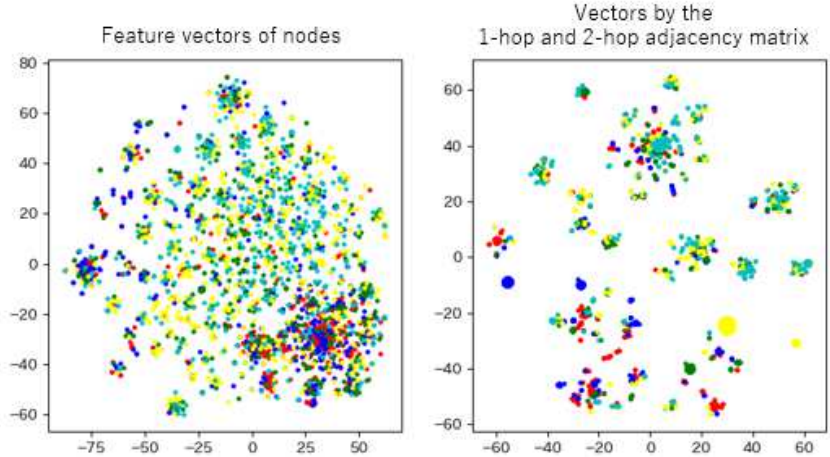}
  \caption{Comparison of usefulness of vectors by visualization through t-SNE}\label{fig:tsnecham}
\end{figure}

In this part of the study, we analyze the usefulness of the structural information of graph data for node classification.
We generate vectors by extracting the rows of the 1-hop and 2-hop adjacency matrices as vectors that contain the structural information of the graph data.
We also use the feature vectors $X_v$ for comparison.
Figure \ref{fig:tsnecham} visualizes these vectors by mapping them to a two-dimensional space through t-distributed stochastic neighbor embedding (t-SNE) \cite{tsne} using the Chameleon \cite{geom} dataset.
The left side of Figure \ref{fig:tsnecham} shows the distribution obtained from the feature vectors, whereas the right side shows the distribution obtained from the vectors containing structural information.
Each point corresponds to one node on the graph and is colored by class.
The points for each class are scattered all over the left side, whereas many clusters for each class are formed on the right side.
This indicates that structural information is more useful  than feature vectors $X_v$ for node classification.

\subsection{Duplicated and Combined Vectors}
\begin{figure}[H]
%\centering
\hspace{-1.0cm}
  \includegraphics[width=13.0cm]{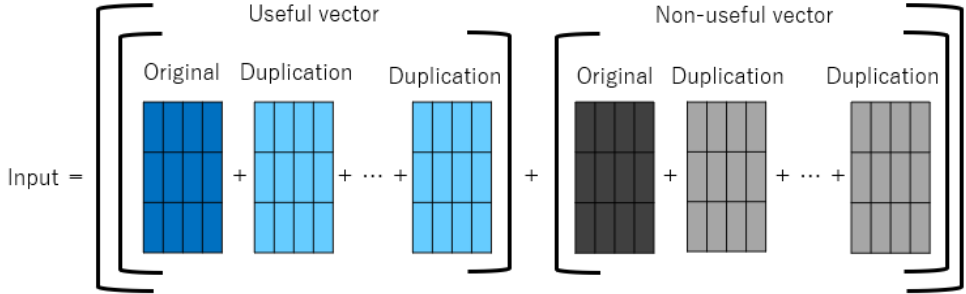}
  \caption{Flow of input vector generation}\label{fig:vecratio}
\end{figure}

\begin{figure}[H]
\centering
  \includegraphics[width=11.0cm]{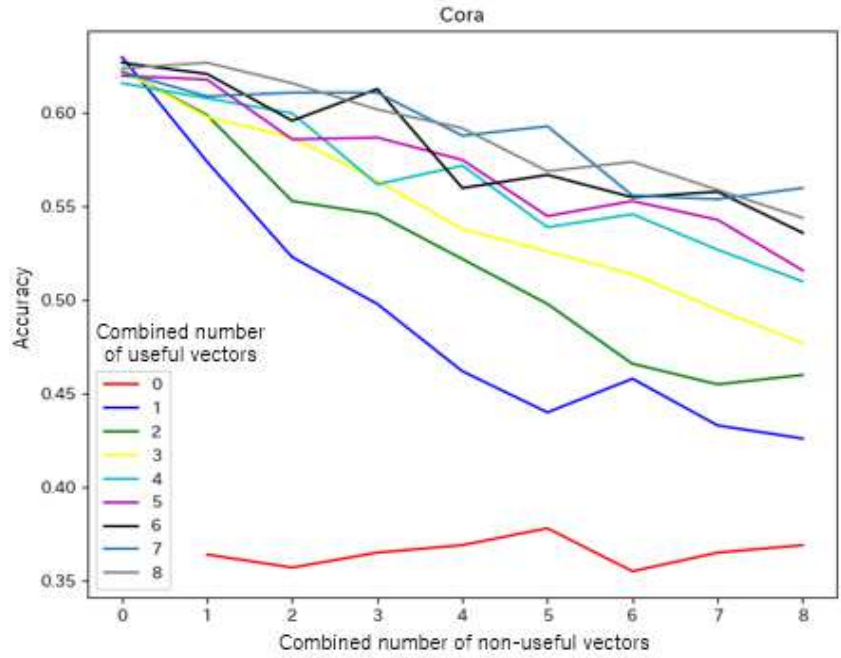}
  \caption{Relationship between combined number and accuracy in GCN}\label{fig:vecratioresult}
\end{figure}

If a vector is duplicated $n$ times, and all of them are combined, we refer to $n$ as the combined number for the vector.
Then, we consider the relationship between the combined number and accuracy of a GNN when a node has both useful and non-useful vectors for node classification.
Let us generate useful 100-dimensional vectors extracted from the feature vectors $X_v$, and non-useful 100-dimensional vectors with random values of 0 and 1.
Then, both vectors are duplicated several times and combined as extended feature vectors to train the GCN model.
Figure \ref{fig:vecratio} shows the flow of the input vector generation.
In the analysis, we observe a transition in the accuracy for each combination of combined numbers.
Figure \ref{fig:vecratioresult} shows the transition in the accuracy when the GCN model is used on the Cora \cite{coracite} dataset.
Because both useful and non-useful vectors are included in the extended feature vector, except in some settings, the amount of information included in the vector does not change even if the ratio of the combined numbers is different.
However, we can observe that the accuracy decreases as the combined number of non-useful vectors increases.
In addition, the decrease in accuracy becomes slower as the combined number of useful vectors increases. 
Therefore, it is important for accuracy improvement to duplicate the vectors and adjust the ratio of the combined number according to the usefulness of the vectors when the nodes have a number of different vectors.
Based on this idea, our MSI-GNN model is designed to duplicate each feature vector and adjust the ratio of the combined numbers.

\section{MSI-GNN}
\begin{figure}[H]
%\centering
\hspace{-0.5cm}
  \includegraphics[width=13.0cm]{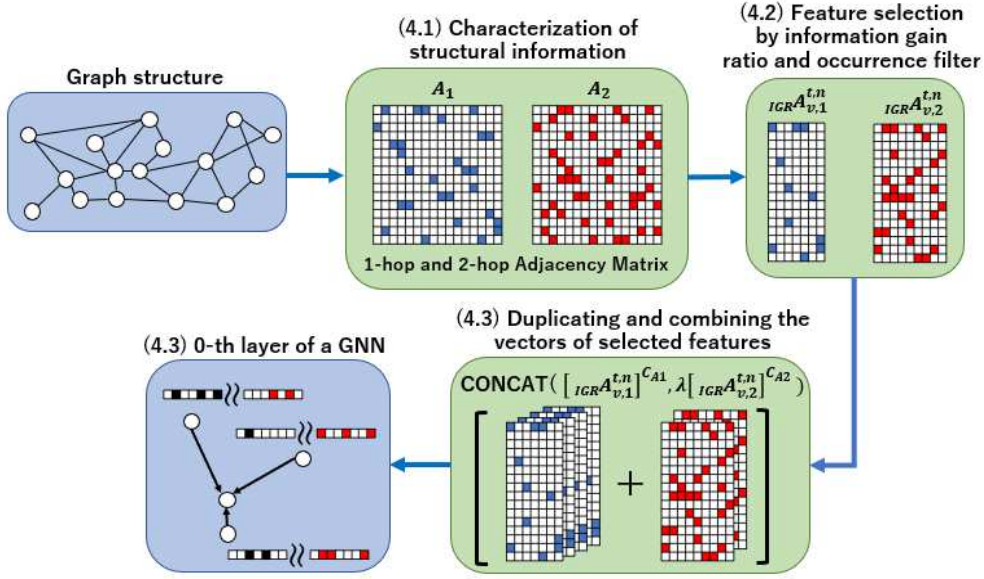}
  \caption{Flow of characterizing structural information of graph data in MSI-GNN}\label{fig:igrgnn}
\end{figure}

In this section, we define MSI-GNN, which characterizes the structural information of graph data.
As shown in Figure \ref{fig:igrgnn}, (4.1) MSI-GNN uses an $i$-hop adjacency matrix to characterize the structural information of graph data.
MSI-GNN then (4.2) selects only useful features according to the information gain ratio and occurrence filter, and (4.3) adjusts the ratio of the combined numbers and incorporates them into the 0-th layer of a GNN.

\subsection{Characterization of Structural Information}
Let $A_i \in \mathbb{Z}^{~V| \times |V|}$ be an $i$-hop adjacency matrix that expresses whether nodes in the graph $G$ are adjacent to each node that is $i$-hops away.
Then, the vector $A_{v, i}$ extracted from $A_i$ as a row vector represents the adjacency status of node $v$ and characterizes the structural information of node $v$.

\subsection{Feature Selection By Information Gain Ratio and Occurrence Filter}
The computational cost of a GNN on large graph data will be much higher when the entire $A_{v, i}$ is used as input. 
Moreover, the structural information for node classification may contain noise.
Therefore, we select only useful structural information according to the information gain ratio.

Let $A_{i}[v,u]$ be a value representing whether nodes $v$ and $u$ are adjacent in the $i$-hop adjacency matrix $A_{i}$.
Additionally, let $V_L^y$ be the set of nodes $v \in V_L$ with label $y$.
Let $V_{L, i}^u$ be the set of labeled nodes $v \in V_L$ that have $u$ as an $i$-hops away node ($A_{i}[v,u] = 1$), and $V_{L, i}^{-u}$ be the set of labeled nodes $v \in V_L$ that do not have $u$ as an $i$-hops away node ($A_{i}[v,u] = 0$).
Then, the amount of information representing the randomness of the data $\infofl{V_L}$ and the amount of information for the data divided by $i$-hops-away node $u$ $\ainfofu{u}{V_L, A_i}$ are defined as follows:

\begin{eqnarray}
\infofl{V_L} & = & \infofr{\vSet{L}}, \nonumber \\
\ainfofu{u}{V_L, A_i} & = & \myainfofr{u}{\vSet{L}}{\vSet{L,i}}. \nonumber
\end{eqnarray}
The information gain ratio $\myigrfl{u}{\vSet{L}, A_i}$ for adjacent node $u$ located $i$-hop away is defined as follows:
\begin{equation}
\myigrfl{u}{\vSet{L}, A_i} = \frac{\infofl{V_L} - \ainfofu{u}{V_L, A_i}}{\mysinfofr{u}{\vSet{L}}{\vSet{L,i}}}. \nonumber
\end{equation}
Next, we sort a node sequence $(u'_1, ..., u'_n)$ in descending order of information gain ratio for all nodes $u_1, ..., u_n \in V$ as follows:
\begin{equation}
\mathrm{ARGSORT}(  \myigrfl{u_1}{\vSet{L}, A_i}, ..., \myigrfl{u_n}{\vSet{L}, A_i} ), \nonumber
\end{equation}
where the $\mathrm{ARGSORT}$ function returns an index sequence (denoted by a node sequence) for the values in descending order.
We then reconstruct the $i$-hop adjacency matrix ${}_{\mathrm{IGR}}A_{i} \in \mathbb{Z}^{|V| \times |V|}$ as follows: 
\begin{equation}
{}_{\mathrm{IGR}}A_{i} = [ A_{u'_1, i}^T, ..., A_{u'_n, i}^T]. \nonumber
\end{equation}
This ${}_{\mathrm{IGR}}A_{i}$ is an $i$-hop adjacency matrix reconstructed in descending order of information gain ratio via calculation of the information gain ratio for all nodes $u$ using $\myigrfl{u}{\vSet{L}, A_i}$.
Then, we extract the nodes with the top $t$ information gain ratios from ${}_{\mathrm{IGR}}A_{i}$ and filter the nodes with unclear usefulness using the occurrence filter.
The matrix ${}_{\mathrm{IGR}}A_{i}^{-}( \in \mathbb{Z}^{|V| \times t})$ containing only this extracted structural information is represented as follows:
\begin{equation}
{}_{\mathrm{IGR}}A_{i}^{-} = \mathrm{FILTER}(\mathrm{SLICE}({}_{\mathrm{IGR}}A_{i}, 1, t), n), \nonumber
\end{equation}
where $\mathrm{SLICE}(a, b, c)$ is an operation for an extraction from the $b$-th column to the $c$-th column of matrix $a$, and FILTER(*, $n$) is an occurrence filter that removes nodes $u$ where $|V_{L,i}^u|$ is less than or equal to $n$ from *.
For example, when $n=1$, nodes that are adjacent to all labeled nodes only once can be removed as noise.

\subsection{MSI-GNN Layer}
We identify the vector ${}_{\mathrm{IGR}}A_{v, i}^{-}$ extracted from the matrix ${}_{\mathrm{IGR}}A_{i}^{-}$ as an IGR-$i$-hop structural vector.
This indicates whether node $v$ is adjacent to the filtered features.
Then, we combine the vectors characterizing the structural information located 1 to $m$-hops away from node $v$ as follows:
\begin{equation}
 \mathrm{CONCAT}\Bigl({}_{\mathrm{IGR}}A_{v, 1}^{-},\ \lambda {}_{\mathrm{IGR}}A_{v, 2}^{-},\ ...,\ \lambda^{m-1} {}_{\mathrm{IGR}}A_{v, m}^{-}\Bigr), \nonumber
\end{equation}
where $\mathrm{CONCAT}(a, b, c, ....)$ is an operation that concatenates vectors $(a,b,c,...)$.
In addition, $\lambda \in [0, 1]$ is a discount coefficient that restricts structural information beyond 2-hops in proportion to the number of hops.

Furthermore, the MSI-layer ${}_{\mathrm{IGR}}S_{v, m}^{ \lambda}$ is obtained when each IGR-$i$-hop structural vector ($1 \leq i \leq m$) is duplicated and combined several times.
\begin{equation}
{}_{\mathrm{IGR}}S_{v, m}^{ \lambda} = \mathrm{CONCAT}\Bigl(  [{}_{\mathrm{IGR}}A_{v, 1}^{-}]^{c_{A1}},\ \lambda [{}_{\mathrm{IGR}}A_{v, 2}^{-}]^{c_{A2}},\ ...,\ \lambda^{m-1} [{}_{\mathrm{IGR}}A_{v, m}^{-}]^{c_{Am}}\Bigr), \nonumber
\end{equation}
where $[a]^{c}$ is a vector combining $c$ vectors $a$ as follows. We refer to $c$ as the combined number.
\begin{equation}
[a]^{c} = {\Big |\Big |}_{k=1}^c a. \nonumber
\end{equation}
Given a feature vector $X_v$, the MSI-layer ${}_{\mathrm{IGR}}S_{v, m}^{ \lambda}$ is defined as follows:
\begin{equation}
{}_{\mathrm{IGR}}S_{v, m}^{\lambda} = \mathrm{CONCAT}\Bigl( [X_v]^{c_{X}}, [{}_{\mathrm{IGR}}A_{v, 1}^{-}]^{c_{A1}}, \lambda [{}_{\mathrm{IGR}}A_{v, 2}^{-}]^{c_{A2}}, ...,\ \lambda^{m-1} [{}_{\mathrm{IGR}}A_{v, m}^{-}]^{c_{Am}}\Bigr) \nonumber
\end{equation}
MSI-layer ${}_{\mathrm{IGR}}S_{v, m}^{ \lambda}$ duplicates and combines each IGR-$i$-hop structural vector and feature vector $X_v$ several times.
In this way, it adjusts the ratio of the combined number of each vector to the learning, thus resulting in a high performance in node classification.

In this paper, we refer to the GNN that incorporates the MSI-layer as MSI-GNN (Multi-duplicated characterization of graph Structures using IGR - GNN).
MSI-GNNs can be applied to various GNNs, such as GCN, GCNII, and H2GCN.
Specifically, each MSI-GNN is defined by the 0-th layer $h_v^{(0)}$ of each GNN as follows:
\begin{equation}
  h_v^{(0)} =
  \begin{cases}
    {}_{\mathrm{IGR}}S_{v, m}^{\lambda} & \text{if GCN is selected,} \\
    \sigma \Bigl( {}_{\mathrm{IGR}}S_{v, m}^{\lambda}\ W_0 + b \Bigr)            & \text{if GCNII is selected,} \\
    \sigma \Bigl( {}_{\mathrm{IGR}}S_{v, m}^{\lambda}\ W_e \Bigr)       & \text{if H2GCN is selected,}
  \end{cases} \nonumber
\end{equation}
where $b$ is the learnable bias weight, and $W_0$ and $W_e$ are the learnable weight matrices.

\section{Related Work}
In this section, we provide a review of related studies on node classification tasks performed on graph data.
There are methods that use graph embedding as an approach to node classification.
Graph embedding converts various pieces of information from graph data into vector representations.
The graph embedding methods DeepWalk \cite{deepwalk} and node2vec \cite{node2vec} learn node embedding representations by sampling neighboring node information and applying it to the Skip-gram algorithm of Word2vec \cite{word2vec}, which was developed for natural language processing (NLP).
DeepWalk uses random sampling by random walk, whereas node2vec uses sampling based on breadth-first search and depth-first search.
Meanwhile, GraphSAGE \cite{sage} learns aggregate functions, which can also generate unknown node embeddings.

A GNN is a message-passing mechanism \cite{gilmer2017neural} that aggregates messages from neighboring nodes and updates its own representation.
The most basic form of GNN is GCN \cite{gcn}, which learns by aggregating the feature vectors of neighboring nodes and averaging them.
Various GNN models based on GCN have since been developed. 
For example, graph attention networks (GAT) \cite{gat} use the attention mechanism to learn different weights for each neighborhood: simple graph convolution (SGC) \cite{sgc} reduces model complexity while maintaining accuracy by removing nonlinear transformation layers in GCN: GCN-LPA \cite{gcnlpa} uses label propagation for regularization: and Geom-GCN \cite{geom} extracts the structural information of graph data from geometric relations in the latent space and uses a two-step aggregation operation.
As a knowledge distillation framework based on GNN, Combination of Parameterized label propagation and Feature transformation (CPF) \cite{cpf} uses a trained GNN as a teacher model and conducts additional learning using multilayer perceptron (MLP) \cite{mlp} and label propagation \cite{lp}.
However, despite the variety of its implementations, GNN has a problem with over-smoothing, in which the representations of the nodes become closer as the number of layers becomes larger.
To solve this problem, the model JKNet \cite{jknet} captures local information by combining representations of hidden layers: DropEdge \cite{dropedge} uses a renormalized graph convolution matrix with randomly removed edges: approximate personalized propagation of neural predictions (APPNP) \cite{appnp} captures a wide range of node information without increasing the number of layers by using Personalized PageRank: and GCNII \cite{gcnii} uses initial residual connection and identity mapping.

Graph data can be categorized based on homophily, which indicates how nodes with the same label tend to exist close together, and heterophily, which indicates how nodes with different labels tend to exist close together.
Conventional GNNs are effective for graph data that exhibit homophily, but do not perform well on graph data that exhibit heterophily.
GNNs that do perform well on graph data that exhibit heterophily include MixHop \cite{mixhop}, which explicitly distinguishes neighborhoods by depth and extracts them: H2GCN \cite{h2}, which further distinguishes between self and neighborhood: CPGNN \cite{cpgnn}, which propagates pre-computed label predictions based on a matrix that holds the probability that each label is connected: FAGCN \cite{fagcn}, which extracts information flexibly by combining high-pass and low-pass filters: and adaptive channel mixing GNN (ACM-GNN) \cite{acm}, which reduces information loss by incorporating a self-filter.

%%%%%%%%%%%%%%%%%%%%%%%%%%%%%%%%%%%%%%%%%%%%%%%%%%%%%%%%%%%%%%%%%%%%%%%%%%
%%%%%%%%%%%%%%%%%%%%%%%%%%%%%%%%%%%%%%%%%%%%%%%%%%%%%%%%%%%%%%%%%%%%%%%%%%
\section{Evaluation}
In this part of study, we conduct experiments to evaluate the performance of MSI-GNN in node classification by comparing its results with those of conventional GNNs.

\subsection{Dataset and Setting}
In this experiment, we use the nine datasets summarized in Table \ref{tab:dataset}.
Each dataset is widely used as a benchmark for GNN.
The details of the datasets are as follows:
\begin{itemize}
\setlength{\itemsep}{-1pt}
\item \textbf{Cora, Citeseer, and Pubmed (Citation dataset)} \cite{pubmed, coracite} are networks that represent the citation relationships between papers. Nodes represent papers, edges represent citation relationships, and labels represent the fields of the papers. The bag-of-words model is used for the feature vector.

\item \textbf{Texas, Wisconsin, and Cornell} \cite{geom} are networks that represent the link relationships between university web-pages. Nodes represent web pages, edges represent link relationships, and labels represent page categories. The bag-of-words model is used for the feature vector.

\item \textbf{Squirrel and Chameleon} \cite{geom} are networks that represent link relationships in Wikipedia that are related to a specific topic. Nodes represent web-pages, edges represent link relationships, and labels are classified from page traffic.

\item \textbf{Actor} \cite{geom} is a network that represents the co-occurrence of actors on Wikipedia. Nodes represent actors, edges represent co-occurrences between pages, and labels are classified from words in Wikipedia pages.
\end{itemize}

We use 10 random splits such that the data are divided into training, test, and validation datasets at proportions of 48\%, 20\%, and 32\%, respectively, for each label based on H2GCN \cite{h2}.

%\vspace{-20pt}
\begin{table}[H]
    \begin{center}
\caption{Dataset overview} \label{tab:dataset}
\hspace{-0.2cm}
  \begin{tabular}{ c | c c c c } \hline
&Node&Edges&Number of Features&Classes\\ \hline
Texas&183&279&1703&5\\
Wisconsin&251&450&1703&5\\
Actor&4600&26659&932&5\\
Squirrel&5201&198353&2089&5\\
Chameleon&2277&31371&2325&5\\
Cornell&183&277&1703&5\\
Cora&2708&5278&1433&7\\
Citeseer&3327&4552&3703&6\\
Pubmed&19717&44327&500&3\\
\hline
\end{tabular}
\end{center}
\end{table}

%%%%%%%%%%%%%%%%%%%%%%%%%%%%%%%%%%%%%%%%%%%%%%%%%%%%%%%%%%%%%%%%%%%%%%
%IGR multi-duplicated structure characterized vector
\subsection{Experimental Setup}
In the experiment, we use MSI-GCN, MSI-H2GCN-$K$ (where $K$ is the number of layers), and MSI-GCNII as MSI-GNN.
We also employ the MSI-layer, which characterizes structural information located 1-hop and 2-hops away, and set the parameter $n$ of the occurrences filter to 1.
The numbers of layers of GCN and MSI-GCN are both set to $k$ = 2, whereas the numbers of dimensions of the hidden layers of GCN and MSI-GCN, and the 0-th layer of H2GCN and MSI-H2GCN-$K$ are all fixed at 64.

The numbers of layers and the hyperparameters $\alpha$ and $\beta$ of GCNII and MSI-GCNII are shown in Table \ref{tab:iiparam} in Appendix \ref{sec:param}.
For the combined number of each vector, we use $c_X \in \{0, 1\}$,  $c_{A1}$, $c_{A2} \in \{0, 1\}$ for citation data, and $c_{A1}$, $c_{A2} \in \{0, 1, 4, 8\}$ for other data.
For the parameter search for the GNN, we fix the information gain ratio parameter $t = 1000$ and the discount coefficient $\lambda = 0.5$.
Then, we use the number of dropout rates $dr \in \{0.0, 0.5\}$, weight decay $wd \in \{ 0.0005, 0.00001 \}$, H2GCN activation function $\sigma \in \{ \mathrm  {ReLU, None} \}$, and all combinations of combined numbers for each vector, and select the best combination of parameters based on the performance on the validation dataset.
Afterward, we search for the parameters for the MSI-GNN using fixed parameter settings for the GNN that maximize the performance on validation dataset when $t = 1000$ and $\lambda = 0.5$.
Then, we use the information gain ratio parameter $t \in \{10, 100, 1000\}$, discount coefficients $\lambda \in \{0.1, 0.5, 1.0\}$, and all combinations of combined numbers for each vector, and select the best combination of parameters based on the performance on the validation dataset.
We use Adam \cite{adam} for the optimizer with a learning rate of 0.01 and number of epochs $e = 1000$.
If the validation loss does not decrease by more than 200 epochs, we stop the training.
The optimal parameter settings determined using the validation data are listed in Table \ref{tab:optparam} in Appendix \ref{sec:param}.

%%%%%%%%%%%%%%%%%%%%%%%%%%%%%%%%%%%%%%%%%%%%%%%%%%%%%%%%%%%%%%%%%%%%%%%

\subsection{Experimental Results}	
\begin{figure}[H]
%\hspace{-0.3cm}
\centering
  \includegraphics[scale=0.6]{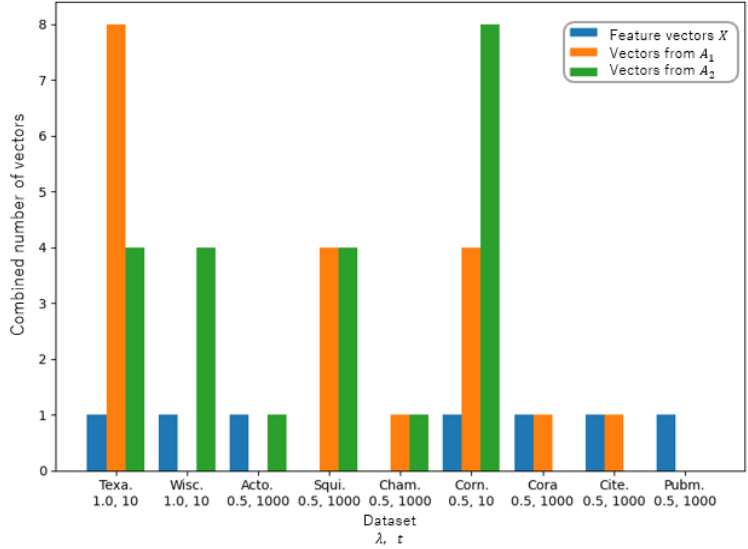}
  \caption{Optimal parameter settings for MSI-H2GCN-2}\label{fig:optparam}
\end{figure}

Table \ref{tab:result} shows the accuracy of each model on each dataset and the average accuracy for all datasets.
The MSI-GNNs are underlined.
The maximum accuracies demonstrated by our own implementation and those reported in other papers are shown in bold.
The rows marked with ``*" indicate the accuracies quoted from papers \cite{gcnii, acm}.
Figure \ref{fig:optparam} shows the optimal parameter settings used in MSI-H2GCN-2, which exhibited the highest average accuracy, as shown in Table \ref{tab:result}.
The observations from these results are summarized as follows.

\begin{itemize}
\item MSI-GCN had a higher accuracy than that of GCN, except on Cornell and Pubm\-ed: MSI-H2GCN had a higher accuracy than that of H2GCN, except on Actor, Cornell, and Pubmed: and MSI-GCNII had a higher accuracy than that of GCNII, except for Actor. This is particularly true for Squirrel, on which accuracy was improved by up to 38.95\%. This indicates that MSI-GNN is an effective  model for node classification compared to conventional GNNs.

\item MSI-H2GCN-2 exhibited the highest accuracy among all models on Squirrel and Chameleon, as shown in Table \ref{tab:result}. According to the optimal parameter settings shown in Figure \ref{fig:optparam}, the optimal settings for these two datasets do not use the feature vectors $X_v$. This indicates that the structural information and adjustment of the combined number in MSI-GNN are beneficial for node classification.

\item MSI-H2GCN-2 had the highest average accuracy, which was 2.46\% higher than that of ACM-GCN, which had the highest average accuracy among the other methods.
\end{itemize}

\begin{table}[H]
    \caption{Experimental Results for each method. ``*" indicates experimental results from reference papers, and ``-" indicates that experiment was not performed. MSI-GNNs are underlined.}
\label{tab:result}
    \hspace{-2.0cm}
  \begin{tabular}{ c || c | c | c | c | c | c | c | c | c || c } \hline
  &Texa.&Wisc.&Acto.&Squi.&Cham.&Corn.&Cora&Cite.&Pubm.&Average acc\\ \hline \hline
\underline{MSI-GCN}&60.81&58.63&29.72&54.85&67.02&55.95&87.34&75.98&87.84&64.24\\
GCN&58.65&57.84&29.66&49.25&66.47&56.49&87.14&75.24&87.84&63.18\\ \hline
\underline{MSI-H2GCN-1}&85.86&84.51&34.78&73.85&79.47&78.92&88.21&76.74&89.43&76.86\\
\underline{MSI-H2GCN-2}&\textbf{86.49}&\textbf{86.08}&35.20&\textbf{74.19}&\textbf{79.85}&80.54&88.09&76.63&89.39&\textbf{77.38}\\
H2GCN-1&83.24&84.90&34.80&33.94&57.11&77.30&87.36&76.66&\textbf{89.44}&69.42\\
H2GCN-2&84.86&85.29&35.20&35.24&57.21&\textbf{81.62}&87.89&76.65&89.39&70.37\\ \hline
\underline{MSI-GCNII}&74.59&76.67&34.91&73.83&79.30&75.68&\textbf{88.33}&\textbf{77.24}&89.30&74.43\\
GCNII&73.51&76.08&\textbf{35.26}&38.13&58.22&74.86&88.25&76.75&89.19&67.81\\
\hline \hline
H2GCN-1*&84.86&86.67&35.86&36.42&57.11&82.16&86.92&77.07&89.40&70.72\\
H2GCN-2*&82.16&85.88&35.62&37.90&59.39&82.16&87.81&76.88&89.59&70.82\\
%GCN(H2GCN論文*)&59.46&59.80&30.26&36.89&59.82&57.03&87.28&76.68&87.38&61.62\\
GCNII*&69.46&74.12&-&-&60.61&62.48&\textbf{88.49}&77.08&89.57&-\\
%Geom-GCN参照データ合ってるか不明*&67.57&64.12&31.63&38.14&60.90&60.81&85.27&\textbf{77.99}&\textbf{90.05}&\\
ACM-GCN*&\textbf{87.84}&\textbf{88.43}&\textbf{36.28}&\textbf{54.40}&\textbf{66.93}&85.14&87.91&\textbf{77.32}&\textbf{90.00}&\textbf{74.92}\\
ACMII-GCN*&86.76&87.45&36.16&51.85&66.91&\textbf{85.95}&88.01&77.15&89.89&74.46\\
%ACM-Snowball-2*&87.57&87.06&\textbf{36.89}&52.50&\textbf{67.08}&85.41&87.42&76.41&89.89&74.47\\
 \end{tabular}
\end{table}

%%%%%%%%%%%%%%%%%%%%%%%%%%%%%%%%%%%%%%%%%%%%%%%%%%%%%%%%%%%%%%%%%%%%%%%%%

\subsection{Usefulness of MSI-Layer}
\begin{figure}[H]
%\hspace{-0.3cm}
\centering
    \includegraphics[width=10.0cm]{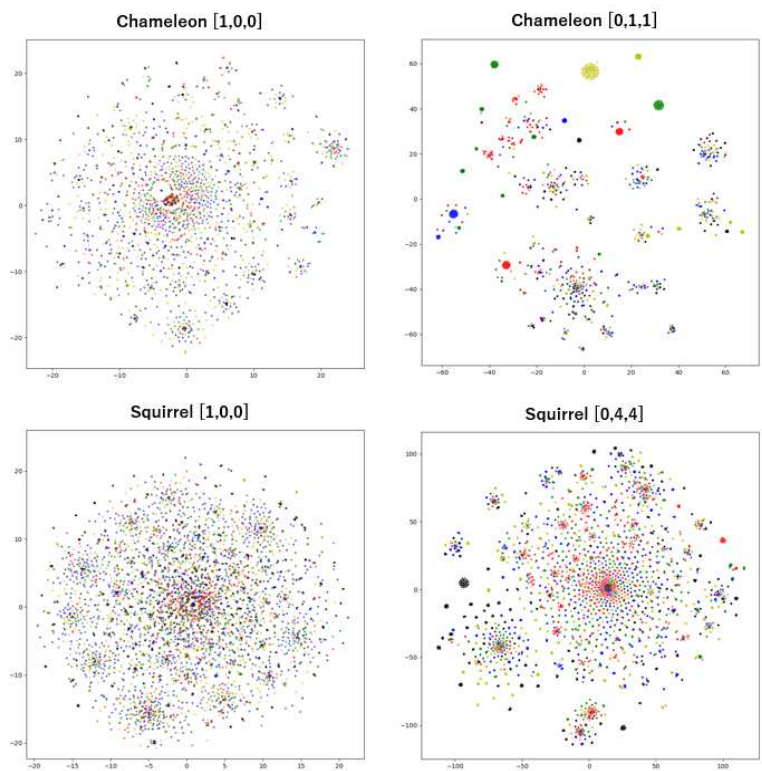}
  \caption{Visualization by t-SNE (input: MSI-layer). Combinations of numbers after each dataset name represent combined number for each vector used ([$c_X, c_{A1}, c_{A2}$]).}\label{fig:feavec}
\end{figure}

\begin{figure}[H]
\centering
    \includegraphics[width=10.0cm]{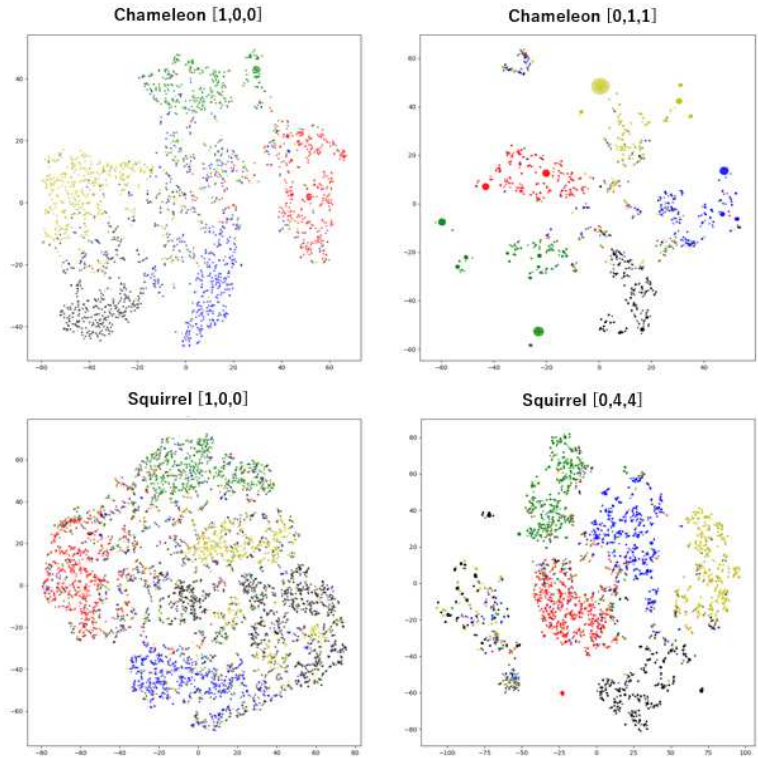}
  \caption{Visualization by t-SNE (input: vectors that represent probabilistic label prediction by output of trained MSI-H2GCN-2). Combinations of numbers after each dataset name represent combined number for each vector used ([$c_X, c_{A1}, c_{A2}$]).}\label{fig:presoftmax}
\end{figure}

We analyze the usefulness of the MSI-layer for node classification.
We visualize the MSI-layers by mapping them to a two-dimensional space using t-SNE \cite{tsne}.
We also use the vectors that represent the probabilistic label prediction based on output from the trained MSI-H2GCN-2 for comparison.
Figure \ref{fig:feavec} shows the distribution from the MSI-layer, whereas Figure \ref{fig:presoftmax} shows the distribution for the output of the trained MSI-H2GCN-2.
We use the Squirrel and Chameleon datasets, on which MSI-H2GCN-2 achieved a high performance.
Each point corresponds to one node and is colored by class.
The combinations of numbers ([0, 4, 4], etc.) after the name of each dataset represents the combined number for each vector ([$c_X, c_{A1}, c_{A2}$]).
In other words, the left side of each figure shows the distribution when only the feature vectors $X_v$ are used, whereas the right side shows the distribution when the optimal parameter settings shown in Figure \ref{fig:optparam} are used.

On the left side of Figure \ref{fig:feavec}, the points are sparsely scattered without any clusters for each class.
By contrast, on the right side of Figure \ref{fig:feavec}, there are several clusters belonging to the same class.
This result indicates that the MSI-layer is useful for node classification, at least for some nodes, compared to the feature vectors $X_v$.
Moreover, on the left side of Figure \ref{fig:presoftmax}, the points in the same class tend to be close.
This is because the vectors of nodes with the same class are trained to obtain a similar vector.
By contrast, on the right side of Figure \ref{fig:presoftmax}, we can observe that the clusters for each class are more clearly clustered.
This indicates that even nodes that have already been predicted to be in their correct classes based on feature vectors $X_v$ can now be predicted more accurately.
From the aforementioned results, we can conclude that the MSI-layer is more useful than the feature vectors $X_v$ for node classification.
In addition, the MSI-GNN can learn without the feature vectors $X_v$ if proper adjustments are made on the ratio of the combined number.
This will be beneficial if the feature vectors are not useful for the node classification.

%%%%%%%%%%%%%%%%%%%%%%%%%%%%%%%%%%%%%%%%%%%%%%%%%%%%%%%%%%%%%%%%%%%%%%%%%%

\subsection{Relationship Between Parameters and Accuracy}
\begin{figure}[H]
%\centering
\hspace{-1.5cm}
  \includegraphics[width=15.0cm]{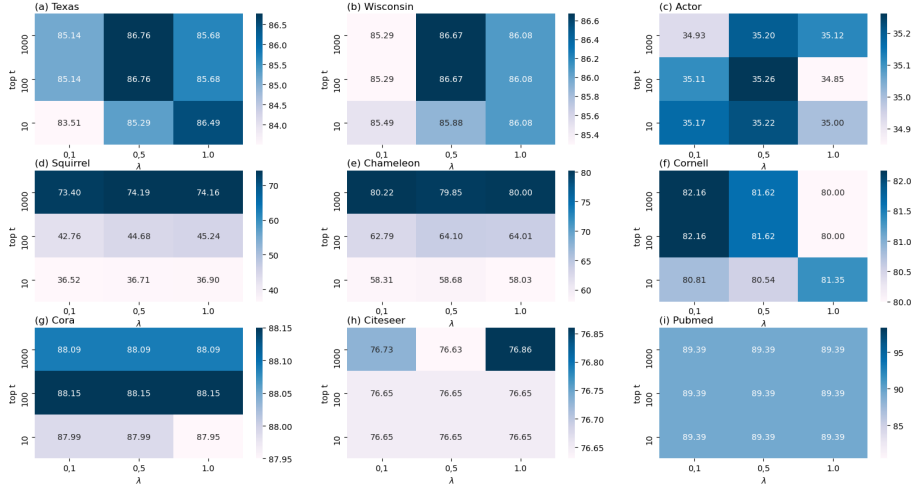}
  \caption{Relationship between parameters and accuracy}\label{fig:heat}
\end{figure}

MSI-GNN utilizes parameters $t$ for extracting useful features based on the information gain ratio and discount coefficient $\lambda$ to restrict deep structural information.
We analyze the relationship between these parameters and accuracy.
Figure \ref{fig:heat} shows the accuracy of MSI-H2GCN-2 for each $t$ and $\lambda$ as a heat map for all datasets.
We use the optimal parameter settings in Figure \ref{fig:optparam}, except for $t$ and $\lambda$.
Because the numbers of nodes of Texas, Wisconsin, and Cornell are less than 1000, the accuracies are the same for both $t = 100$ and 1000.
In addition, the accuracy on Pubmed does not change because using only the feature vectors $X_v$ is the optimal parameter setting.
With regard to the discount coefficient, the accuracies on Texas, Wisconsin, and Cornell tend to increase when $\lambda = 0.1$ or 0.5.
This indicates that it is effective to restrict structural information beyond 2-hops.
On the other hand, with regard to the information gain ratio, the accuracy is maximized when the top 1000 features of Squirrel and Chameleon are used.
By contrast, the accuracy drops by more than 10\% when only the top 100 features are used.
Furthermore, for all of the graph data, the accuracy tends to decrease when only the top 10 features are used.
This indicates that even features with a low information gain ratio may be useful for node classification.

%%%%%%%%%%%%%%%%%%%%%%%%%%%%%%%%%%%%%%%%%%%%%%%%%%%%%%%%%%%%%%%%%%%%%%%%
	
\section{Conclusion}
In this paper, we propose MSI-GNN, which characterizes the structural information of graph data.
MSI-GNN selects only useful features for node classification based on the information gain ratio and occurrence filter.
Furthermore, MSI-GNN realizes flexible and powerful learning by adaptively duplicating and combining structural vectors and feature vectors.
In the evaluation experiments using multiple datasets, we applied MSI-GNN to GCN, H2GCN, and GCNII to improve their performance.
In particular, MSI-H2GCN-2 achieved the highest average accuracy.

In future work, we will consider a novel method for selecting more useful features from the adjacency matrix.
We will also need to apply MSI-GNN to state-of-the-art GNN models.
This may result in greater performance gains.
Additionally, our MSI-layer can be applied to other tasks such as graph classification and graph embedding.

%%%%%%%%%%%%%%%%%%%%%%%%%%%%%%%%%%%%%%%%%%%%%%%%%%%%%%%%%%%%%%%%%%%%%%%%
%%%%%%%%%%%%%%%%%%%%%%%%%%%%%%%%%%%%%%%%%%%%%%%%%%%%%%%%%%%%%%%%%%%%%%%%

\small
\bibliographystyle{junsrt}
\bibliography{firstinitial_gakkai}

\appendix
\section{Parameter Settings}\label{sec:param}
Table \ref{tab:iiparam} shows the settings for the dimensions of the hidden layers, numbers of layers, and hyperparameters of GCNII and MSI-GCNII in the experiment outlined in Table \ref{tab:result}. We refer to GCNII \cite{gcnii} to set these parameters.
Table \ref{tab:optparam} lists the optimal parameter settings determined by the performance of the validation data in the experiments outlined in Table \ref{tab:result}. Parameters not used by each model are marked with ``-".

\begin{table}[H]
\centering
    \caption{Parameter settings in GCNII and MSI-GCNII}
\label{tab:iiparam}
  \begin{tabular}{ c | c c c c } \hline
Dataset&Dimension of hidden layers&Number of layers&$\alpha$&$\beta$ \\ \hline \hline
Texas&64&32&0.5&1.5\\
Wisconsin&64&16&0.5&1.0\\
Actor&64&16&0.5&1.0\\
Squirrel&64&16&0.5&1.0\\
Chameleon&64&8&0.2&1.5\\
Cornell&64&16&0.5&1.0\\
Cora&64&64&0.2&0.5\\
Citeseer&64&64&0.5&0.5\\
Pubmed&64&64&0.1&0.5\\ \hline
 \end{tabular}
\end{table}

\begin{table}[H]
%\centering
    \caption{Optimal parameter settings}
\label{tab:optparam}
    \hspace{-2.7cm}
\scriptsize
  \begin{tabular}{ c | c | c  c  c c c  c  c  c | c } \hline
Dataset&Model&Regularization weight&Dropout&Activation function&Discount coefficient&$t$&$C_X$&$C_{A1}$&$C_{A2}$&Accuracy of test data\\ \hline \hline
Texas
&GCN&5e-4&0.0&-&-&-&-&-&-&58.65\\
&MSI-GCN&5e-4&0.0&-&1.0&1000&1&8&4&60.81\\
&H2GCN-1&5e-4&0.5&ReLU&-&-&-&-&-&83.24\\
&H2GCN-2&5e-4&0.5&ReLU&-&-&-&-&-&84.86\\
&MSI-H2GCN-1&5e-4&0.5&ReLU&1.0&1000&1&1&4&85.86\\
&MSI-H2GCN-2&5e-4&0.5&ReLU&1.0&10&1&8&4&86.49\\
&GCNII&5e-4&0.0&-&-&-&-&-&-&73.51\\
&MSI-GCNII&5e-4&0.5&-&1.0&1000&1&8&8&74.59\\
\hline
Wisconsin
&GCN&5e-4&0.0&-&-&-&-&-&-&57.84\\
&MSI-GCN&5e-4&0.0&-&0.5&1000&1&8&8&58.63\\
&H2GCN-1&5e-4&0.5&ReLU&-&-&-&-&-&84.90\\
&H2GCN-2&5e-4&0.5&ReLU&-&-&-&-&-&85.29\\
&MSI-H2GCN-1&5e-4&0.5&ReLU&1.0&10&1&0&1&84.51\\
&MSI-H2GCN-2&5e-4&0.5&ReLU&1.0&10&1&0&4&86.08\\
&GCNII&5e-4&0.5&-&-&-&-&-&-&76.08\\
&MSI-GCNII&5e-4&0.5&-&1.0&1000&1&1&1&76.67\\
\hline
Actor
&GCN&5e-4&0.5&-&-&-&-&-&-&29.66\\
&MSI-GCN&5e-4&0.5&-&0.1&10&1&1&8&29.72\\
&H2GCN-1&5e-4&0.5&ReLU&-&-&-&-&-&34.80\\
&H2GCN-2&5e-4&0.5&None&-&-&-&-&-&35.20\\
&MSI-H2GCN-1&5e-4&0.5&ReLU&0.1&10&1&8&1&34.78\\
&MSI-H2GCN-2&5e-4&0.5&None&0.5&1000&1&0&0&35.20\\
&GCNII&1e-5&0.5&-&-&-&-&-&-&35.26\\
&MSI-GCNII&1e-5&0.5&-&0.1&10&1&0&8&34.91\\
\hline
Squirrel
&GCN&1e-5&0.0&-&-&-&-&-&-&47.25\\
&MSI-GCN&1e-5&0.0&-&0.5&1000&0&1&0&54.85\\
&H2GCN-1&5e-4&0.0&ReLU&-&-&-&-&-&33.94\\
&H2GCN-2&5e-4&0.0&ReLU&-&-&-&-&-&35.24\\
&MSI-H2GCN-1&5e-4&0.0&ReLU&1.0&1000&0&1&1&73.85\\
&MSI-H2GCN-2&5e-4&0.0&ReLU&0.5&1000&0&4&4&74.19\\
&GCNII&1e-5&0.5&-&-&-&-&-&-&38.13\\
&MSI-GCNII&5e-4&0.0&-&1.0&1000&0&8&4&73.83\\
\hline
Chameleon
&GCN&1e-5&0.0&-&-&-&-&-&-&66.47\\
&MSI-GCN&1e-5&0.0&-&0.1&1000&1&0&1&67.02\\
&H2GCN-1&1e-5&0.5&ReLU&-&-&-&-&-&57.11\\
&H2GCN-2&1e-5&0.5&ReLU&-&-&-&-&-&57.21\\
&MSI-H2GCN-1&5e-4&0.5&ReLU&0.5&1000&0&1&1&79.47\\
&MSI-H2GCN-2&5e-4&0.5&ReLU&0.5&1000&0&1&1&79.85\\
&GCNII&5e-4&0.5&-&-&-&-&-&-&58.22\\
&MSI-GCNII&1e-5&0.5&-&0.5&1000&0&8&8&79.30\\
\hline
Cornell
&GCN&5e-4&0.0&-&-&-&-&-&-&56.49\\
&MSI-GCN&5e-4&0.5&-&0.1&10&1&1&4&55.95\\
&H2GCN-1&5e-4&0.5&ReLU&-&-&-&-&-&77.30\\
&H2GCN-2&5e-4&0.5&ReLU&-&-&-&-&-&81.62\\
&MSI-H2GCN-1&5e-4&0.5&ReLU&0.1&1000&1&1&8&78.92\\
&MSI-H2GCN-2&5e-4&0.5&ReLU&0.5&10&1&4&8&80.54\\
&GCNII&5e-4&0.5&-&-&-&-&-&-&74.86\\
&MSI-GCNII&5e-4&0.5&-&0.1&1000&1&1&4&75.68\\
\hline
Cora
&GCN&5e-4&0.5&-&-&-&-&-&-&87.14\\
&MSI-GCN&5e-4&0.5&-&0.1&100&1&1&1&87.34\\
&H2GCN-1&5e-4&0.5&ReLU&-&-&-&-&-&87.36\\
&H2GCN-2&5e-4&0.5&None&-&-&-&-&-&87.89\\
&MSI-H2GCN-1&5e-4&0.5&ReLU&0.5&1000&1&1&1&88.21\\
&MSI-H2GCN-2&5e-4&0.5&ReLU&0.5&1000&1&1&0&88.09\\
&GCNII&5e-4&0.5&-&-&-&-&-&-&88.25\\
&MSI-GCNII&5e-4&0.5&-&0.1&100&1&1&1&88.33\\
\hline
Citeseer
&GCN&5e-4&0.5&-&-&-&-&-&-&75.24\\
&MSI-GCN&5e-4&0.5&-&0.1&10&1&1&1&75.98\\
&H2GCN-1&5e-4&0.5&ReLU&-&-&-&-&-&76.66\\
&H2GCN-2&5e-4&0.5&ReLU&-&-&-&-&-&76.65\\
&MSI-H2GCN-1&5e-4&0.5&ReLU&0.5&1000&1&1&0&76.74\\
&MSI-H2GCN-2&5e-4&0.5&None&0.5&1000&1&1&0&76.63\\
&GCNII&5e-4&0.5&-&-&-&-&-&-&76.75\\
&MSI-GCNII&5e-4&0.5&-&0.5&1000&1&1&0&77.24\\
\hline
Pubmed
&GCN&1e-5&0.5&-&-&-&-&-&-&87.84\\
&MSI-GCN&1e-5&0.5&-&1000&0.5&1&0&0&87.84\\
&H2GCN-1&1e-5&0.5&ReLU&-&-&-&-&-&89.44\\
&H2GCN-2&1e-5&0.5&ReLU&-&-&-&-&-&89.39\\
&MSI-H2GCN-1&1e-5&0.5&ReLU&0.1&10&1&0&1&89.43\\
&MSI-H2GCN-2&1e-5&0.5&ReLU&0.5&1000&1&0&0&89.39\\
&GCNII&1e-5&0.5&-&-&-&-&-&-&89.19\\
&MSI-GCNII&1e-5&0.5&-&0.1&10&1&1&1&89.30\\
\hline
 \end{tabular}
\end{table}

\end{document}